\documentclass[arabic,11pt]{article}
\usepackage{arabtex}
\usepackage{utf8}
\usepackage{acl2014}
\usepackage{times}
\usepackage{url}
\usepackage{latexsym}
\usepackage{graphicx}
\usepackage{amsmath}

\setcode{utf8}

\title{Arabic Spelling Correction using Supervised Learning}

\author{Youssef Hassan \\
  	Dept Computer Engineering\\
Cairo University \\
	Giza, Egypt \\
  {\small{\tt youssefhassan13@gmail.com}} \\\And
  Mohamed Aly \\
  Dept Computer Engineering \\
  Cairo University \\
  Giza, Egypt \\
  \small{{\tt mohamed@mohamedaly.info}} \\\And
  Amir Atiya \\
  Dept Computer Engineering \\
  Cairo University \\
  Giza, Egypt \\
  \small{{\tt amir@alumni.caltech.edu}} \\
}

\date{}

\begin{document}
\maketitle
\begin{abstract}
  In this work,\ we address the problem of spelling correction in the Arabic language utilizing the new corpus provided by QALB (Qatar Arabic Language Bank) project which is an annotated corpus of sentences with errors and their corrections. The corpus contains edit, add before, split, merge, add after, move and other error types.\ We are concerned with the first four error types as they contribute more than 90\% of the spelling errors in the corpus. The proposed system has many models to address each error type on its own and then integrating all the models to provide an efficient and robust system that achieves an overall recall of 0.59, precision of 0.58 and F1 score of 0.58 including all the error types on the development set.\ Our system participated in the QALB 2014 shared task "Automatic Arabic Error Correction" and achieved an F1 score of 0.6, earning the sixth place out of nine participants.
\end{abstract}

\section{Introduction}

The Arabic language is a highly inflected natural language that has an enormous number of possible words \cite{Othman}.\ And although it is the native language of over 300 million people, it suffers from the lack of useful resources as opposed to other languages, specially English and until now there are no systems that cover the wide range of possible spelling errors.\ Fortunately the QALB corpus \cite{ZAGHOUANI14.956.L14-1721} will help enrich the resources for Arabic language generally and the spelling correction specifically by providing an annotated corpus with corrected sentences from user comments, native student essays, non-native data and machine translation data.\ In this work, we are trying to use this corpus to build an error correction system that can cover a range of spelling errors.

This paper is a system description paper that is submitted in the EMNLP 2014 conference shared task "Automatic Arabic Error Correction" \cite{Mohit-etal-qalb2014} in the Arabic NLP workshop.\ The challenges that faced us while working on this system was the shortage of contribution in the area of spelling correction in the Arabic language.\ But hopefully the papers and the work in this shared task specifically and in the workshop generally will enrich this area and flourish it.

Our system targets four types of spelling errors, edit errors, add before errors, merge errors and split errors. For each error type, A model is built to correct erroneous words detected by the error detection technique. Edit errors and add before errors are corrected using classifiers with contextual features, while the merge and split errors are corrected by inserting or omitting a space between words and choosing the best candidate based on the language model score of each candidate.

The rest of this paper is structured as follows. In section 2, we give a brief background on related work in spelling correction. In section 3, we introduce our system for spelling correction with the description of the efficient models used in the system.\ In section 4, we list some experimental results on the development set.\ In section 5, we give some concluding remarks.

\section{Related Work}
The work in the field of spelling correction in the Arabic language is not yet mature and no system achieved a great error correction efficiency. Even Microsoft Word, the most widely used Arabic spelling correction system, does not achieve good results.\ Our work was inspired by a number of papers.\ \cite{SHAALAN12.603} addressed the problem of Arabic Word Generation for spell checking and they produced an open source and large coverage word list for Arabic containing 9 million fully inflected surface words and applied language models and Noisy Channel Model and knowledge-based rules for error correction. This word list is used in our work besides using language models and Noisy Channel Model. 

\cite{5461784} proposed another system  for cases in which the candidate generation using edit algorithm only was not enough, in which candidates were generated based on transformation rules and errors are detected using BAMA (Buckwalter Arabic Morphological Analyzer)\cite{buckwalter}.

\cite{Khalifa_arabicdiscourse} proposed a system for text segmentation. The system discriminates between waw wasl and waw fasl, and depending on this it can predict if the sentence to be segmented at this position or not, they claim that they achieved 97.95\% accuracy. The features used in this work inspired us with the add before errors correction.

\cite{Schaback_multi-levelfeature} proposed a system for the English spelling correction, that is addressing the edit errors on various levels: on the phonetic level using \textit{Soundex} algorithm, on the character level using edit algorithm with one operation away, on the word level using bigram language model, on the syntactic level using collocation model to determine how fit the candidate is in this position and on the semantic level using co-occurrence model to determine how likely a candidate occurs within the given context, using all the models output of candidate word as features and using SVM model to classify the candidates, they claim reaching recall ranging from 90\% for first candidate and 97\% for all five candidates presented and outperforming MS Word, Aspell, Hunspell, FST and Google. 

\section{Proposed System}
We propose a system for detecting and correcting various spelling errors, including edit, split, merge, and add before errors.\ The system consists of two steps: error detection and error correction.\ Each word is tested for correctness. If the word is deemed incorrect, it is passed to the correction step, otherwise it remains unchanged. The correction step contains specific handling for each type of error, as detailed in subsection \ref{ssect:Edit Errors Correction}.

\subsection{Resources}
{\bf Dictionary}: Arabic wordlist for spell checking\footnote{\url{http://sourceforge.net/projects/arabic-wordlist/}} is a free dictionary containing 9 million Arabic words.\ The words are automatically generated from the AraComLex\footnote{\url{http://aracomlex.sourceforge.net/}} open-source finite state transducer.

The dictionary is used in the generation of candidates and using a special version of MADAMIRA\footnote{MADAMIRA-release-20140702-1.0} \cite{Pasha} created for the QALB shared task using a morphological database based on BAMA 1.2.1\footnote{AraMorph 1.2.1 - \url{http://sourceforge.net/projects/aramorph/}} \cite{buckwalter}. Features are extracted for each word of the dictionary to help in the proposed system in order that each candidate has features just like the words in the corpus.

{\bf Stoplist}: Using stop words list available on sourceforge.net\footnote{\url{http://sourceforge.net/projects/arabicstopwords/}}. This is used in the collocation algorithm described later.

{\bf Language Model}:\ We use SRILM \cite{Stolcke} to build a language model using the Ajdir Corpora\footnote{\url{http://aracorpus.e3rab.com/argistestsrv.nmsu.edu/AraCorpus/}} as a corpus with the vocabulary from the dictionary stated above.\ We train a language model containing unigrams, bigrams, and trigrams using modified Kneser-Ney smoothing \cite{James:2000:MKS:891196}.

{\bf QALB Corpus}: QALB shared task offers a new corpus for spelling correction.\ The corpus contains a large dataset of manually corrected Arabic sentences.\ Using this corpus, we were able to implement a spelling correction system that targets the most frequently occurring error types which are \textbf{(a) edit errors} where a word is replaced by another word, \textbf{(b) add before errors} where a word was removed, \textbf{(c) merge errors} where a space was inserted mistakenly and finally \textbf{(d) split errors} where a space was removed mistakenly. The corpus provided also has three other error types but they occur much less frequently happen which are \textbf{(e) add after errors} which is like the add before but the token removed should be put after the word, \textbf{(f) move errors} where a word should be moved to other place within the sentence and \textbf{(g) other errors} where any other error that does not lie in the six others is labeled by it.

\subsection{Error Detection}
The training set, development set and test set provided by QALB project come with the "columns file" and contains very helpful features generated by MADAMIRA. Using the Buckwalter morphological analysis \cite{buckwalter} feature, we determine if a word is correct or not. If the word has no analysis, we consider the word as \textit{incorrect} and pass it through the correction process.

\subsection{Edit Errors Correction}
\label{ssect:Edit Errors Correction}
The edit errors has the highest portion of total errors in the corpus. It amounts to more than 55\% of the total errors.\ To correct this type of errors, we train a classifier with features like the error model probability, collocation and co-occurrence as follows:

{\bf Undiacriticized word preprocessed}: Utilizing the MADAMIRA features of each word, the undiacriticized word fixes some errors like hamzas, the pair of haa and taa marboutah and the pair of yaa and alif maqsoura. 

We apply some preprocessing on the undiacriticized word to make it more useful and fix the issues associated with it. For example we remove the incorrect redundant characters from the word e.g  (\<الرجاااال> $\rightarrow$  \<الرجال>, AlrjAAAAl $\rightarrow$ AlrjAl). We also replace the Roman punctuation marks by the Arabic ones e.g (? $\rightarrow$ \<؟>).

{\bf Language Model}: For each candidate, A unigram, bigram and trigram values from the language model trained are retrieved. In addition to a feature that is the product of the unigram, bigram and trigram values.

{\bf Likelihood Model}: The likelihood model is trained by iterating over the training sentences counting the occurrences of each edit with the characters being edited and the type of edit. The output of this is called a confusion matrix.  

The candidate score is based on the Noisy Channel Model \cite{Kernighan:1990:SCP:997939.997975} which is the multiplication of probabilty of the proposed edit using the confusion matrix trained which is called the error model, and the language model score of that word.\ The language model used is unigram, bigram and trigram with equal weights. Add-1 smoothing is used for both models in the counts.
						\begin{center} $Score = p(x|w) . p(w)$ \end{center}
where $x$ is the wrong word and $w$ is the candidate correction.

For substitution edit candidates, we give higher score for substitution of a character that is close on the keyboard or the substitution pair belongs to the same group of letter groups \cite{SHAALAN12.603} by multiplying the score by a constant greater than one.
\begin{arabtext}
(آ‬, ‫إ‬, ‫أ‬,‫ ا‬), (ي, ‫ن‬,‫ ث‬, ‫ت‬,‫ ب‬), (خ‬, ‫ح‬, ‫ج‬), (ذ‬, ‫د‬), (ز‬, ‫ر‬), (ش‬, ‫س‬), (ض‬, ‫ص‬), (ظ‬, ‫ط‬), (غ‬,‫ ع‬), (ق‬, ‫ف‬), (ة‬،‫ ه‬), (‫ؤ‬, ‫و‬), (‫ى‬,‫ ي‬).
\end{arabtext}
 ($|$‬, $‫<‬$, ‫$>$‬,‫ A‬), (y, ‫n‬,‫ v‬, ‫t‬,‫ b‬), (x‬, ‫H‬, ‫j‬), (*‬, ‫d‬), (z‬, ‫r‬), (\$‬, ‫s‬), (D‬, ‫S‬), (Z‬, ‫T‬), (g‬,‫ E‬), (q‬, ‫f‬), (p‬،‫ h‬), (‫\&‬, ‫w‬), (‫Y‬,‫ y‬)

For each candidate , the likelihood score is computed and added to the feature vector of the candidate.

{\bf Collocation}: The collocation model targets the likelihood of the candidate inside the sentence. This is done using the lemma of the word and the POS tags of words in the sentence.

We use the algorithm in \cite{Schaback_multi-levelfeature} for training the collocation model. Specifically, by retrieving the 5,000 most occurring lemmas in the training corpus and put it in list $L$.\ For each lemma in $L$, three lists are created, each record in the list is a sequence of three POS tags around the target lemma. For training, we shift a window of three POS tags over the training sentence. If a lemma belongs to $L$, we add the surrounding POS tags to the equivalent list of the target lemma depending on the position of the target lemma within the three POS tags. 

Given a misspelled word in a sentence, for each candidate correction, if it is in the $L$ list, we count the number of occurrences of the surrounding POS tags in each list of the three depending on the position of of the candidate.

The three likelihoods are stored in the feature vector of the candidate in addition to the product of them.

{\bf Co-occurrence}: Co-occurrence is used to measure how likely a word fits inside a context. Where $L$ is the same list of most frequent lemmata from collocation. 

We use the co-occurrence algorithm in \cite{Schaback_multi-levelfeature}. Before training the model, we transform each word of our training sentence into its lemma form and remove stop-words. For example, consider the original text:

\begin{center}\<حيث لأفرق بين الاستعمار والحكومة الحالية بما أنها >

Hyv l$>$frq byn AlAstEmAr wAlHkwmp AlHAlyp bmA $>$nhA \end{center}

After removing stop-words and replacing the remaining words by their lemma form we end up with:

\begin{center}\<أفرق استعمار حكومة حالي>

$>$frq AstEmAr Hkwmp HAly \end{center}
which forms $C$.

From that $C$, we get all  lemmata that appear in the radius of 10 words around the target lemma $b$ where $b$ belongs to $L$.\ We count the number of occurrences of each lemma in that context $C$.

By using the above model, three distances are calculated for target lemma $b$: $d_1$, the ratio of actually found context words in $C$ and possibly findable context words. This describes how similar the trained context and the given context are for candidate $b$; $d_2$ considers how significant the found context lemmata are by summing the normalized frequencies of the context lemmata.\ As a third feature; $d_3(b)$ that simply measures how big the vector space model for lemma $b$ is.

For each candidate, the model is applied and the three distances are calculated and added to the feature vector of that candidate.

{\bf The Classifier}:
After generating the candidate corrections within 1 and 2 edit operations (insert, delete, replace and transpose) distance measured by Levenshtein distance \cite{levenshtein1966bcc}, we run them through a Naive-Bayes classifier using python NLTK's implementation to find out which one is the most likely to be the correction for the incorrect word.

The classifier is trained using the training set provided by QALB project. For each edit correction in the training set, all candidates are generated for the incorrect word and a feature vector (as shown in table\ref{tab:edit_features}) is calculated using the techniques aforementioned. If the candidate is the correct one, the label for the training feature vector is \textitً{correct} else it is incorrect.

\begin{table}[h]
\caption{The feature set used by the edit errors classifier.\\} 
\centering 
	\begin{tabular}{|l | l |}
		\hline
		\it{Feature name} \\ \hline
		Likelihood model probability\\ \hline
		unigram probability \\ \hline
		previous bigram probability  \\ \hline
		next bigram probability  \\ \hline
		trigram probability  \\ \hline
		language model product \\ \hline
		collocation left \\ \hline
		collocation right \\ \hline
		collocation mid \\ \hline
		collocation product \\ \hline
		cooccurrence distance 1 \\ \hline
		cooccurrence distance 2 \\ \hline
		cooccurrence distance 3 \\ \hline
		previous gender \\ \hline
		previous number \\ \hline
		next gender \\ \hline
		next number \\ \hline
		
	\end{tabular}
\label{tab:edit_features}
\end{table}

Then using the trained classifier, the same is done on the development set or the test set where we replace the incorrect word with the word suggested by the classifier.

\subsection{Add before Errors Correction}
The add before errors are mostly punctuation errors. A classifier is trained on the QALB training corpus. A classifier is implemented with contextual features $C$. $C$ is a 4-gram around the token being investigated. Each word of these four has the two features: The token itself and Part-of-speech tag and for the next word  only pregloss because if the word's pregloss is "and" it is more probable that a new sentence began. Those features are available thanks to MADAMIRA features provided with the corpus and the generated for dictionary words.

The classifier is trained on the QALB training set.\ We iterate over all the training sentences word by word and getting the aforementioned features (as shown in table \ref{tab:add_before_features}) and label the training with the added before token if there was a matching add before correction for this word or the label will be an empty string. 

For applying the model, the same is done on the QALB development sentences after removing all punctuations as they are probably not correct and the output of the classifier is either empty or suggested token to add before current word.

\begin{table}[h]
\caption{The feature set used by the add before errors classifier.\\} 
\centering 
	\begin{tabular}{|l | l |}
		\hline
		\it{Feature name} \\ \hline
		before previous word\\ \hline
		before previous word POS tag  \\ \hline
		previous word\\ \hline
		previous word POS tag  \\ \hline
		next word\\ \hline
		next word POS tag  \\ \hline
		next word pregloss\\ \hline
		after next word\\ \hline
		after next POS tag  \\ \hline
	\end{tabular}
\label{tab:add_before_features}
\end{table}

\subsection{Merge Errors Correction}
The merge errors occurs due to the insertion of a space between two words by mistake.\ The approach is simply trying to attach every word with its successor word and checking if it is a valid Arabic word and rank it with the language model score.

\subsection{Split Errors Correction}
The split errors occurs due to the deletion of a space between two words. The approach is simply getting all the valid partitions of the word and try to correct both partitions and give them a rank using the language model score.\ The partition is at least two characters long. 

\section{Experimental Results}
In order to know the contribution of each error type models to the overall system performance, we adopted an incremental approach of the models.\ We implemented the system using python\footnote{https://www.python.org/} and NLTK\footnote{http://www.nltk.org/} \cite{loper} toolkit. The models are trained on the QALB corpus training set and the results are obtained by applying the trained models on the development set. Our goal was to achieve high recall but without losing too much precision. The models were evaluated using M2 scorer \cite{Dahlmeier:2012:BEG:2382029.2382118}.

First, we start with only the preprocessed undiacriticized word, then we added our edit error classifier. Adding the add before classifier was a great addition to the system as the system was able to increase the number of corrected errors significantly, notably the add before classifier proposed too many incorrect suggestions that decreased the precision.\ Then we added the merging correction technique. Finally we added the split error correction technique.\ The system corrects 9860 errors versus 16659 golden error corrections and proposed 17057 correction resulting in the final system recall of 0.5919, precision of 0.5781 and F1 score of 0.5849. Details are shown in Table \ref{tab:Results}.

\begin{table}[h]
\small
\caption{The incremental results after adding each error type model and applying them on the development set.} 
\centering 
	\begin{tabular}{|l | l | l | l | l |}
		\hline
		Model name & Recall & Precision  & F1 score\\ \hline
		Undiacriticized & 0.32 & 0.833  & 0.4715\\ \hline
		+ Edit & 0.3515 & 0.7930  & 0.5723\\ \hline
		+ Add before & 0.5476 & 0.5658  & 0.5567\\ \hline
		+ Merge & 0.5855 & 0.5816  & 0.5836 \\ \hline
		+ Split & \textbf{0.5919} & \textbf{0.5781}  & \textbf{0.5849}\\ \hline
	\end{tabular}
\label{tab:Results}
\end{table}

We tried other combinations of the models by removing one or more of the components to get the best results possible. Noting that all the systems results are using the undiacriticized word. Details are shown in Table \ref{tab:Results2}
\begin{table}[h]
\small
\caption{The results of some combinations of the models and applying them on the development set. The models are abbreviated as Edit E, Merge M, Split S, and Add before A.} 
\centering 
	\begin{tabular}{|l | l | l | l | l |}
		\hline
		Model name & Precision & Recall & F1 score\\ \hline
		M Only & 0.8441 & 0.3724 & 0.5167\\ \hline
		S Only & 0.7838 & 0.338 & 0.5167 \\ \hline
		A Only & 0.6008 & 0.4887 & 0.539\\ \hline
		E Only & 0.8143 & 0.3472 & 0.4868\\ \hline
		M \& S & 0.8121 & 0.3814 & 0.5191\\ \hline
		E \& S & 0.62 & 0.3542 & 0.4508\\ \hline
		M \& E & 0.6184 & 0.5403 & 0.5767\\ \hline
		S \& M \& A	&0.6114	&0.5396	&0.5733\\ \hline
		M \& E \& A	&0.6186	&0.5404	&0.5768	\\ \hline
		E \& S \& A	&0.5955	&0.507	&0.5477	\\ \hline
		E \& S \& M &0.6477&0.3969	&0.4922	\\ \hline
		\textbf{E \& S \& M \& A} & \textbf{0.5919} & \textbf{0.5781}  & \textbf{0.5849}\\ \hline
	\end{tabular}
\label{tab:Results2}
\end{table}
\section{Conclusion and Future Work}

We propose an all-in-one system for error detection and correction.\ The system addresses four types of spelling errors (edit, add before, merge and split errors).\ The system achieved promising results by successfully getting corrections for about 60\% of the spelling errors in the development set. Also, There is still a big room for improvements in all types of error correction models.

We are planning to improve the current system by incorporating more intelligent techniques and models for split and merge.\ Also, the add before classifier needs much work to improve the coverage as the errors are mostly missing punctuation marks. For the edit classifier, real-word errors need to be addressed.

\bibliographystyle{acl}
\bibliography{acl2014}

\begin{thebibliography}{}

\bibitem[\protect\citename{Buckwalter}2002]{buckwalter}
Tim Buckwalter.
\newblock 2002.
\newblock Buckwalter arabic morphological analyzer version 1.0.
\newblock November.

\bibitem[\protect\citename{Dahlmeier and
  Ng}2012]{Dahlmeier:2012:BEG:2382029.2382118}
Daniel Dahlmeier and Hwee~Tou Ng.
\newblock 2012.
\newblock Better evaluation for grammatical error correction.
\newblock In {\em Proceedings of the 2012 Conference of the North American
  Chapter of the Association for Computational Linguistics: Human Language
  Technologies}, NAACL HLT '12, pages 568--572, Stroudsburg, PA, USA.
  Association for Computational Linguistics.

\bibitem[\protect\citename{James}2000]{James:2000:MKS:891196}
Frankie James.
\newblock 2000.
\newblock Modified kneser-ney smoothing of n-gram models.
\newblock RIACS.

\bibitem[\protect\citename{Kernighan \bgroup et al.\egroup
  }1990]{Kernighan:1990:SCP:997939.997975}
Mark~D. Kernighan, Kenneth~W. Church, and William~A. Gale.
\newblock 1990.
\newblock A spelling correction program based on a noisy channel model.
\newblock In {\em Proceedings of the 13th Conference on Computational
  Linguistics - Volume 2}, COLING '90, pages 205--210, Stroudsburg, PA, USA.
  Association for Computational Linguistics.

\bibitem[\protect\citename{Khalifa \bgroup et al.\egroup
  }2011]{Khalifa_arabicdiscourse}
Iraky Khalifa, Zakareya~Al Feki, and Abdelfatah Farawila.
\newblock 2011.
\newblock Arabic discourse segmentation based on rhetorical methods.

\bibitem[\protect\citename{Levenshtein}1966]{levenshtein1966bcc}
VI~Levenshtein.
\newblock 1966.
\newblock {Binary Codes Capable of Correcting Deletions, Insertions and
  Reversals}.
\newblock volume~10, page 707.

\bibitem[\protect\citename{Loper and Bird}2002]{loper}
Edward Loper and Steven Bird.
\newblock 2002.
\newblock {NLTK: The Natural Language Toolkit}.

\bibitem[\protect\citename{Mohit \bgroup et al.\egroup
  }2014]{Mohit-etal-qalb2014}
Behrang Mohit, Alla Rozovskaya, Nizar Habash, Wajdi Zaghouani, and Ossama
  Obeid.
\newblock 2014.
\newblock {The First QALB Shared Task on Automatic Text Correction for Arabic}.
\newblock In {\em Proceedings of EMNLP Workshop on Arabic Natural Language
  Processing}, Doha, Qatar, October.

\bibitem[\protect\citename{Othman \bgroup et al.\egroup }2003]{Othman}
Eman Othman, Khaled Shaalan, and Ahmed Rafea.
\newblock 2003.
\newblock A chart parser for analyzing modern standard arabic sentence.
\newblock In {\em To appear in In proceedings of the MT Summit IX Workshop on
  Machine Translation for Semitic Languages: Issues and Approaches}, Louisiana,
  U.S.A.

\bibitem[\protect\citename{Pasha \bgroup et al.\egroup }2014]{Pasha}
Pasha, Arfath, Mohamed Al-Badrashiny, Mona Diab, Ahmed~El Kholy, Ramy Eskander,
  Nizar Habash, Manoj Pooleery, Owen Rambow, and Ryan~M. Roth.
\newblock 2014.
\newblock Madamira: A fast, comprehensive tool for morphological analysis and
  disambiguation of arabic.
\newblock In {\em In Proceedings of the Language Resources and Evaluation
  Conference (LREC)}, Reykjavik, Iceland.

\bibitem[\protect\citename{Schaback}2007]{Schaback_multi-levelfeature}
Johannes Schaback.
\newblock 2007.
\newblock Multi-level feature extraction for spelling correction.
\newblock Hyderabad, India.

\bibitem[\protect\citename{Shaalan \bgroup et al.\egroup }2010]{5461784}
K.~Shaalan, R.~Aref, and A~Fahmy.
\newblock 2010.
\newblock An approach for analyzing and correcting spelling errors for
  non-native arabic learners.
\newblock In {\em Informatics and Systems (INFOS), 2010 The 7th International
  Conference on}, pages 1--7, March.

\bibitem[\protect\citename{Shaalan \bgroup et al.\egroup }2012]{SHAALAN12.603}
Khaled Shaalan, Mohammed Attia, Pavel Pecina, Younes Samih, and Josef van
  Genabith.
\newblock 2012.
\newblock Arabic word generation and modelling for spell checking.
\newblock In Nicoletta Calzolari~(Conference Chair), Khalid Choukri, Thierry
  Declerck, Mehmet~Uğur Doğan, Bente Maegaard, Joseph Mariani, Asuncion
  Moreno, Jan Odijk, and Stelios Piperidis, editors, {\em Proceedings of the
  Eight International Conference on Language Resources and Evaluation
  (LREC'12)}, Istanbul, Turkey, may. European Language Resources Association
  (ELRA).

\bibitem[\protect\citename{Stolcke}2002]{Stolcke}
A.~Stolcke.
\newblock 2002.
\newblock Srilm -- an extensible language modeling toolkit.
\newblock In {\em Proc. Intl. Conf. on Spoken Language Processing},
  Denver,U.S.A.

\bibitem[\protect\citename{Zaghouani \bgroup et al.\egroup
  }2014]{ZAGHOUANI14.956.L14-1721}
Wajdi Zaghouani, Behrang Mohit, Nizar Habash, Ossama Obeid, Nadi Tomeh, Alla
  Rozovskaya, Noura Farra, Sarah Alkuhlani, and Kemal Oflazer.
\newblock 2014.
\newblock Large scale arabic error annotation: Guidelines and framework.
\newblock In {\em Proceedings of the Ninth International Conference on Language
  Resources and Evaluation (LREC'14)}, Reykjavik, Iceland, May. European
  Language Resources Association (ELRA).

\end{thebibliography}

\end{document}